%% file: root.tex
\documentclass[letterpaper, 10pt, conference]{ieeeconf}      

\IEEEoverridecommandlockouts                              

\usepackage{graphicx}
\usepackage{efbox}

\usepackage{subcaption}
\usepackage{amsmath}
\usepackage{nicematrix} 
\usepackage{multirow}
\usepackage{comment}
\usepackage{float} 
\usepackage{dblfloatfix}

\usepackage{tikz, pgfplots, fix-cm}
\usetikzlibrary{plotmarks}
\usepackage{graphicx,booktabs,array}

\usepackage{tikz}
\newcommand*\circled[1]{\tikz[baseline=(char.base)]{
            \node[shape=circle,draw,inner sep=1pt] (char) {#1};}}

\usepackage{amssymb}
\usepackage{pifont}
\usepackage[hyphens]{url}
\usepackage{hyperref}

\def\secref#1{Sec.~\ref{#1}}
\def\figref#1{Fig.~\ref{#1}}
\def\tabref#1{Tab.~\ref{#1}}
\def\eqref#1{Eq.~(\ref{#1})}

\efboxsetup{linecolor=black,linewidth=2pt}

\title{\LARGE \bf Towards Autonomous Visual Navigation in Arable Fields}

\author{Alireza Ahmadi, Michael Halstead, and Chris McCool
	\thanks{All authors are with the University of Bonn, Bonn 53115 Germany. 
			{\tt\small \{alireza.ahmadi, michael.halstead, cmccool\}@uni-bonn.de}}%
}

\newcommand{\hyperfootnote}[1][]{\def\ArgI\hyperfootnoteRelay}

\newcommand\etal{\emph{et al.}}

\pgfplotsset{compat=1.16}

\begin{document}
\maketitle
\thispagestyle{empty}
\pagestyle{empty}

\begin{abstract}


Autonomous navigation of a robot in agricultural
fields is essential for every task from crop monitoring
to weed management and fertilizer application. Many current
approaches rely on accurate GPS, however, such technology
is expensive and can be impacted by lack of coverage. 
As such, autonomous navigation through sensors that
can interpret their environment (such as cameras) is important
to achieve the goal of autonomy in agriculture.
In this paper, we introduce a purely vision-based navigation
scheme that is able to reliably guide the robot through
row-crop fields using computer vision and signal processing techniques without manual intervention. 
Independent of any global localization or mapping, this approach is able to accurately follow the crop-rows and switch between the rows, only using onboard cameras. 
The proposed navigation scheme can be deployed in a wide range of fields with different canopy shapes in various growth stages, creating a crop agnostic navigation approach. 
This was completed under various illumination conditions using simulated and real fields where we achieve an average navigation accuracy of $3.82cm$ with minimal human intervention (hyper-parameter tuning) on BonnBot-I.



\end{abstract}

\section{Introduction}
\label{sec:indroduction}
\input{contribs/introduction}

\section{Related Work}
\label{sec:relatedworks}
\input{contribs/relatedworks}

\section{Autonomous Navigation in Row-Crop Fields}
\label{sec:autoNav}
\input{contribs/auto_navigation}

\section{Vision-Based Guidance in Farming Fields}
\label{sec:vison_based_guidance}
\input{contribs/vision_based_guidance}

\section{Experimental Evaluations}
\label{sec:exp}
\input{contribs/exp}

\section{Conclusion}
\label{sec:conc}
In this paper, we presented a novel approach to enable autonomous navigation in row-crop fields empowering precision farming and crop monitoring tasks.
This approach exploits the crop-row structure using only the local observation from the on-board cameras without requiring any global or local position awareness.
To achieve this, we have proposed a novel multi-crop-row detection strategy that can deal with cluttered and weedy scenes. 
We also proposed a novel lane switching strategy which enables BonnBot-I to switch to a new lane independent of any global positioning system or human intervention.
We evaluated our approach on BonnBot-I on up to five crop types (with varying canopy shapes) in real field conditions and three challenging simulated fields achieving an average navigation accuracy of $3.82cm$ in real fields.
Future work could explore alternative approaches to detecting individual plants (crop/weed semantic segmentation) and consider how global positioning could augment the robustness of the current system.



\bibliography{references}
\bibliographystyle{IEEEtran}

\end{document}

%% file: contribs/introduction.tex


Novel agricultural robotic technologies need to ensure they meet the needs of the key stakeholder, farmers.
Usually, this means being cost effective both as a platform and as a labor (or mechanical) replacement technology.
However, another critical element is that the technology can be deployed to a variety of fields and environmental conditions with minimal intervention (supervision) from the farmer.
In particular, navigating through the field is central to the real-world deployment of agricultural robotics.

Hence, automated field navigation for a robot is essential for every task.
For autonomous navigation it is common for platforms to use precise real-time kinematic (RTK) GNSS as it is being used in fully controlled and engineered agricultural sites where they heavily rely on structural information~\cite{Garford, Bawden17_1}.
But, robotic technologies will never be guaranteed accurate GPS in every field, first due to its expense and second because of the coverage.
An example of this is that currently most fields are seeded using traditional methods, not auto-seeding geo-referenced systems, which creates a gap between GPS capabilities and farming requirements~\cite{Garford}.
As such, utilizing GPS technology in unregistered fields increases the risk of damaging crops~\cite{bakkenrobot}.
In a worst-case scenario GPS can fail, therefore, generally applicable crop-row following techniques are required.

\begin{figure}[t!]
    \vspace{+2mm}
	\centering
	\includegraphics[width=\linewidth]{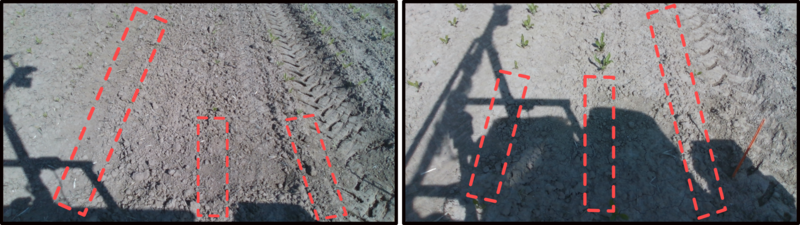}
	\caption{Example images where crops have yet to germinate leaving large gaps in a row, indicated by red boxes.}
	\label{fig:cropRowGaps}
	\vspace{-6mm}
\end{figure}

In our prior work~\cite{ahmadi2020visual} we introduced a GPS independent technique that was able to traverse a single crop-row and switch between adjacent rows.
This system was deployed on a Husky and tested using a single artificial cropping environment (including artificial plants).
Furthermore, the prior work was only applicable to navigating along a single crop-row and did not take advantage of the overall planting structure in the field.
For instance, row-crops are planted in parallel lines and this structure can be exploited to overcome issues such as a lack of germination, which can leave gaps in a crop-row, creating navigation issues for single crop-row techniques. 
Such issues can lead to unrecoverable failure cases for a single-crop-row navigation system, see~\figref{fig:cropRowGaps}.

\begin{figure}[t]
	\centering
	\vspace{2mm}
	\includegraphics[width=0.9\linewidth]{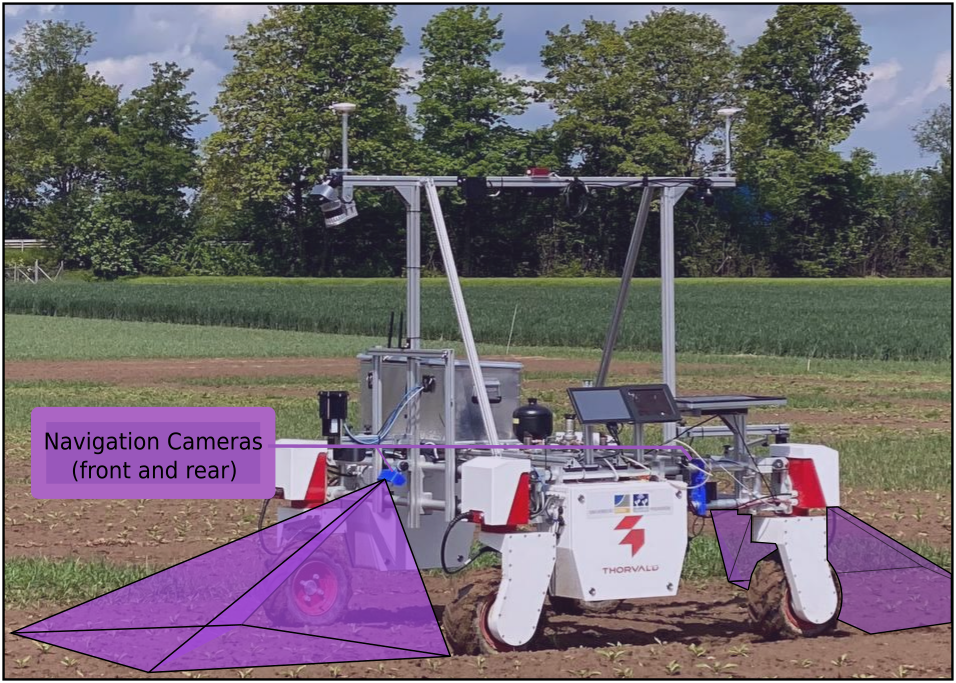}
	\caption{BonnBot-I following lanes of crops using two symmetrically mounted monocular camera in front and back.}
	\label{fig:motivation}
	\vspace{-6mm}
\end{figure}

In this paper we introduce a novel crop-agnostic vision based navigation scheme as illustrated in~\figref{fig:motivation}.
We greatly extend our prior work~\cite{ahmadi2020visual} and use information from multiple crop-rows to navigate a weeding robotic platform in five different real fields under various weather conditions. 
The proposed approach takes advantage of planting schemes (standardized distances between rows of crops), ensures minimum damage to crops and persistent coverage throughout the field.
The proposed approach relies on well-known signal processing and computer vision approaches, however, their combination is unique and their robustness has been demonstrated by being applied to five crop types on a robot in the field.
We consider this approach as an extra navigation controller modality for achieving more reliable traverses in challenging real fields conditions and not denying the potential of benefiting other technologies like GPS and odometry which represents a complete system.
To achieve this we make the following novel contributions:
%
\begin{itemize}
    \item a robust multi-crop-row detection strategy, capable of dealing with cluttered and weedy scenes for a variety of crop types; and
    \item a multiple crop-row following approach which enables autonomous guidance in real-fields; and
    \item a multi-crop-row lane switching strategy which enables BonnBot-I to switch to a new lane
    independent of any global positioning system; and 
    \item deployed and evaluated on a real-sized weed management robot in simulation and the field for five crop types and (with different shapes: straight and curved crop-rows); and 
    \item releasing a novel crop-row dataset covering five crop types with various growth stages under varying illumination conditions.
\end{itemize}

%% file: contribs/relatedworks.tex


Autonomous agricultural robots could improve productivity~\cite{van2002autonomous, bonadies2019overview}, enable targeted field interventions~\cite{perez2014co} and facilitate crop monitoring~\cite{bayati2018mobile}.
For uptake of these platforms they should be deploy-able in a variety of scenarios including different cultivars, crop-row structures, and seeding patterns~\cite{utstumo2018robotic}.
A key enabling technology is reliable navigation through the whole field~\cite{aastrand2005mec}.
One potential solution to the navigation problem is to make use of the Global Navigation Satellite System (GNSS).
Such an approach has been used for both agricultural machinery~\cite{thuilot2002automatic} and robotic platforms~\cite{Bawden17_1}.
The downside of this approach is that it relies on an expensive sensor and suffers from limitations such as possible GNSS outages and reliance on geo-referenced auto-seeding.
Thus, crop-based navigation techniques leveraging the field structure were investigated 
for autonomous guidance~\cite{billingsley1997successful,ahmadi2020visual} and in-field interventions~\cite{aastrand2005vision}.



In an attempt to use the structure of a field,  Barawid~\etal~\cite{barawid2007development} investigated LiDAR based orchards navigation system a similar strategy was used in a simulated environment by~\cite{malavazi2018lidar} for traversing row-crop fields.
In addition, Winterhalter~\etal~\cite{winterhalter2018crop} proposed sensor-independent feature representation from different sensor modalities and detects crop rows. 
They used LiDAR and camera images to extract single lines in a row-crop field which were spaced equidistantly.
While these approaches enable side-applications such as obstacle avoidance, frame drift in self-similar environments can cause issues~\cite{bakken2019end}, including crop damage.

To avoid crop damage through GNSS or LiDAR failures, RGB based navigation approaches directly exploit the available visual information.
These techniques can vary significantly in terms of cost, algorithm simplicity, and availability.
Classical machine vision approaches detect crop-rows with juvenile crops~\cite{aastrand2005vision} or detect crop-rows under challenging illumination conditions~\cite{sogaard2003determination}.
%
Other classical approaches include crop stem locations from multi-spectral images~\cite{haug2014plant}, and plant stem emerging point (PSEP) using hand-crafted features~\cite{midtiby2012estimating}.
While these approaches are generally real-time and can, to varying degrees, navigate a lane they do require hand selected features which reduces the learning abilities of the techniques.


Kraemer~\etal~\cite{kraemer2017plants} used a deep learning approach to reconcile the PSEP features by exploiting the likelihood maps of deep neural networks (DNN).
Also utilising DNNs,~\cite{ponnambalam2020autonomous} proposed a convolutional neural network (CNN) for strawberry crop-row detection with accurate navigation.
Lin~\etal~\cite{lin2019development} also showcase the potential of CNNs to reliably navigate a tea field by classifying tea rows.
These approaches are often more accurate than their traditional counterparts for detecting or segmenting specific plants.
However, in contrast to traditional approaches, CNNs require a significant amount of labeled data and more computational resources both for training and inference, while being less dynamic in nature and requiring further tuning in different conditions.~\cite{de2021towards}.


To perform vision based navigation two common approaches exist: proportional-integral-derivative (PID) or visual servoing. 
Billingsley~\etal~\cite{billingsley1997successful} extracted the row of plants using a Hough transform, and used a PID controller for navigation through a sugar-beet field.
The visual-servoing~\cite{espiau1992new} technique was also exploited for autonomous car guidance in urban areas~\cite{de2014visual} by road lane following with obstacle avoidance using monocular camera.
These methods regularize the controlled agents motion within a loop directly based on the current visual features.

The technique proposed in this paper draws inspiration from our previous work~\cite{ahmadi2020visual} where we are able to both navigate a single crop-row and switch lanes at the end.
However, this approach was only tested in a single artificially created row-crop field, without considering real open-field challenges like (different crop types, illumination variation, appearance of weeds and uneven distribution of plants in rows).
We propose a real-field applicable method to automatically detect the number of crop rows.
Multi-crop-row switching is then enabled by being able to correctly identify new crops as the robot moves across the field.
As real-time performance is important our methods rely on traditional machine vision techniques which we deploy on a real agricultural robot (BonnBot-I) to autonomously navigate in real fields with five different crops and three simulated fields. 
We also made its implementation publicly available \footnote{\url{https://github.com/Agricultural-Robotics-Bonn/visual-multi-crop-row-navigation}}.

%% file: contribs/auto_navigation.tex
Our navigation strategy is influenced by the robot that we use.
In this work we use a retrofitted Thorvald platform~\cite{grimstad2017thorvald} which we refer to as BonnBot-I.
Below we provide, in brief, the specifications of the BonnBot-I platform and then present the high level guidance strategy of our proposed navigation scheme.

\subsection{BonnBot-I Platform}
\label{subsec:platform}
\input{contribs/phenobotPlatform}

\begin{figure}[!t]
	\centering
	\vspace{2mm}
	\includegraphics[width=0.8\linewidth]{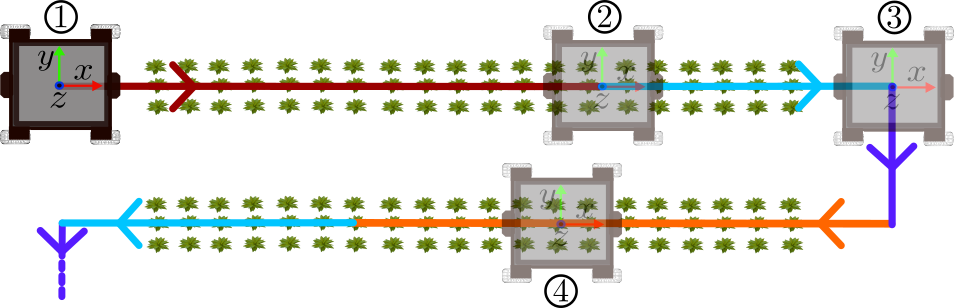}
	\caption{In-field navigation strategy, (1) following crop-rows with front camera, (2) exiting crop-rows using back camera, (3) switching to the next crop-rows, and (4) following new crop-rows with back camera.}
	\label{fig:strategy}
	\vspace{-6mm}
\end{figure}

\subsection{In Field Guidance Strategy}
\label{subsec:guidanceStrategy}
\input{contribs/strategy}

%% file: contribs/phenobotPlatform.tex
BonnBot-I is a prototype crop monitoring and weeding platform.
%
The platform is a lightweight Thorvald system which has been adapted for arable farming and phenotyping fields in Europe.
The European phenotypic regulation pattern ($35\,cm$ or $55\,cm$ between crop-rows) leads to either two or three crop-rows in each lane with a total width of $1.25m$.
For this, the width of the platform was set to $1.4m$ wheel-centre-to-wheel-centre with a vertical clearance of $0.86m$ to ensure the platform could operate in the field during the early growth cycle of most plants.
The length of the robot was set to $1.4m$ ensuring there is room for all the weeding tools, and has a maximum speed of $1.5\,m/s$.
BonnBot-I has multiple sensors, of interest for navigation a GNSS (with an IMU) and two RGB-D cameras which are mounted equally distant from the center of rotation of the robot symmetrically (in front and back) illustrated in~\figref{fig:motivation}.

%% file: contribs/strategy.tex
A benefit of crop-rows is that they are generally planted in consistent long parallel structures.
A downside to this parallel structure is that there is no connection between them.
Therefore, the platform needs to not only follow the crop-row without damaging the crop but also autonomously switch between them.
To achieve multi-crop-row following we employ a similar high-level algorithm to our previous work in~\cite{ahmadi2020visual} for a single crop-row.
\figref{fig:strategy} outlines our multi-crop-row following strategy.
Starting in a corner of the field, the platform autonomously follows the current set of crop-rows (a lane) using vision based navigation techniques~$\circled{1}$ until the front facing camera detects the end of the current lane.
The rear camera then guides the robot to the exit point, end of the lane, actively guiding the robot at all times~$\circled{2}$. 
Using the omni-directional capability, the robot then switches to the next set of  crop-rows~$\circled{3}$ to be traversed.
The benefits of the omni-directional platform prevails here as we can directly navigate to the new lane without turning around~$\circled{4}$, this also outlines the benefit of symmetrically mounted sensors at the front and rear.
In the next section we describe the vision-based crop-row following and crop-row switching algorithms.

%% file: contribs/vision_based_guidance.tex
We propose a multi-crop-row following approach that can successfully guide a real-sized agricultural robot in row-crop fields of five different crops.
In achieving this we have developed a multi-crop-row detection approach described in~\secref{subsec:crop_row_extraction}, integrate information from multiple detected crop-rows to perform visual-servoing based crop-row following in~\secref{subsubsec:vs_nav}, and present a multi-crop-row switching approach in~\secref{subsubsec:switching}.

\subsection{Multi-Crop-Row Detection}
\label{subsec:crop_row_extraction}
\input{contribs/crop_row_extraction}

\subsection{Visual-Servoing Based Crop-Row Following}
\label{subsubsec:vs_nav}

\input{contribs/vs}

\subsection{Multi-Crop-Row Switching}
\label{subsubsec:switching}
\input{contribs/switching}

%% file: contribs/crop_row_extraction.tex

The first step in a successful vision system capable of traversing a field is the assumption that crop-rows are planted in a parallel fashion. 
To have a completely crop agnostic navigation system the varying distances between the rows for the different crops is an important element. 
Therefore, it is imperative to have a system that can detect the number of crop-rows before instantiating the navigation algorithm. 



We perform crop-row detection by employing a novel sliding window based approach.
This extracts the location of the prominent crop-rows while being robust to the appearance of weeds between them.
Our detection approach consists of three steps. 
First, we perform vegetation segmentation followed by connected components operations to find individual regions (plants) and their center points.
This allows us to remain agnostic to the crop that has been planted.
Second, we automatically detect the number of crop-rows by employing an estimate of the moving-variance which we use to describe the field structure.
Finally, we track the detected crop-rows by centering a parallelogram on each row while the robot is traversing the lane.
We detail each of these steps below.
\begin{figure}[t!]
    \vspace{3mm}    
    \begin{subfigure}[b]{42mm}
        \includegraphics[height=28.8mm,width=43mm]{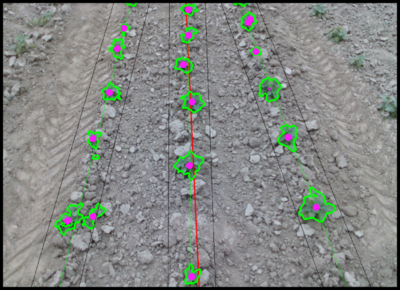}
        \caption{Lemon-balm}
        \label{fig:lemon_balm_graphic}
    \end{subfigure}    
    \vspace{4pt}    
    \begin{subfigure}[b]{42mm}
        \includegraphics[width=43mm]{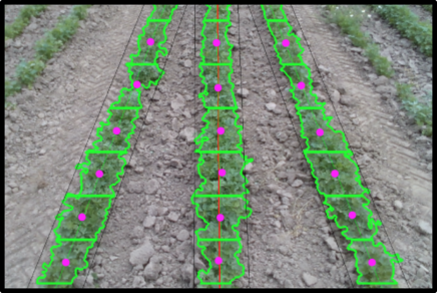}
        \caption{Coriander}
        \label{fig:coriander_graphics}
    \end{subfigure} 
    
    \vspace{-2mm} 
    \caption{
    The vegetation segmentation (in green) with plant boundaries (in bold green) and the resultant plant centers (magenta dots).
    In (a) individual plants are easy to see and (b) is a case where crop boundaries have to be estimated.} 
    
\label{fig:observation_space}
\vspace{-6mm} 
\end{figure}

\subsubsection{Vegetation Mask and Crop Center Extraction}
\label{subsec:vegetation_idx}
In the first stage, we summarize a row by the position of the individual plants along it.
Each plant is represented by its center point.
We obtain this by first 
computing the vegetation mask of the input RGB image using the excess green index (ExG)~\cite{woebbecke1995asae}.
To separate foreground and background pixels in the image based on ExG we employ Otsu's method~\cite{reimann2005background} which obviates the need for manual tuning of a threshold.
Then, each connected component in the vegetation mask is converted to an object 
of interest (plant) with the unique center point obtained from the center of mass. 

One issue associated with this technique occurs when multiple ``plants'' are absorbed into a single large region most often occurring with bushy plants. 
A single region representing an entire crop-row negatively impacts later stages such as line fitting.
To reconcile this we divide contours into smaller subsections if they exceed a predefined maximum height, depicted in~\figref{fig:observation_space}.
Ultimately, this step allows us to cater for a larger variety of canopy types.

\subsubsection{Detecting Individual crop-rows}
\label{subsec:crop_row_detection}

\begin{figure}[t!]
	\centering
	\vspace{3mm}
	\includegraphics[height=6cm]{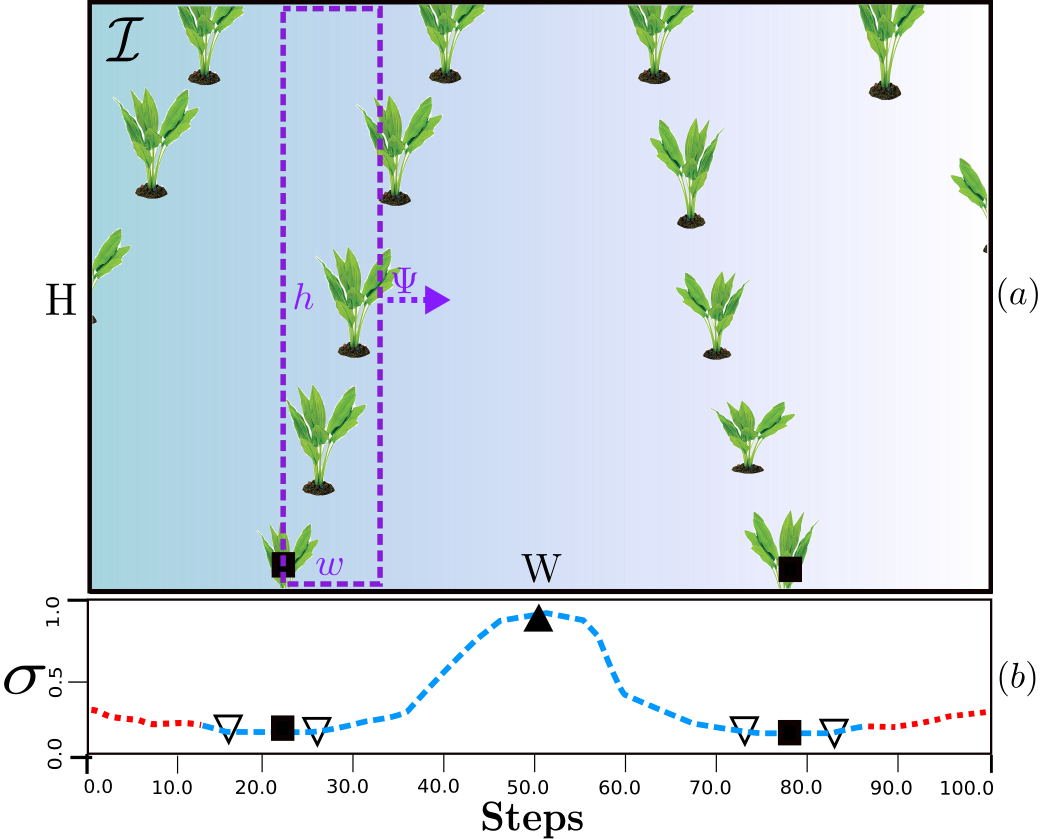}
	\caption{The sliding window $\Psi$ is applied progressively. 
	The moving-variance of the estimated line angles are used to represent the field structure.
	The peaks ($\blacktriangle$) and troughs ($\triangledown$) from the field structure are used to find the crop-rows and the center of the between crop-rows respectively.
	The weighted average of multiple troughs leads to the final trough $\blacksquare$.}
	\label{fig:movar_graph}
	\vspace{-6mm}
\end{figure}

\figref{fig:movar_graph} illustrates the crop-row detection algorithm, 
First, a sliding window $\Psi$ scans the image from left to right with a stride of~$\mathbf{S}$.
The size of the sliding window ($w$ and $h$) and the stride $\mathbf{S}$ are set to ensure a large overlap between adjacent steps in the scans.
For the $n$-th sliding window $\Psi_{n}$ we compute a line ($L_{n}$) based on the crop centers inside the sliding window using least-squares method.
We then find the intersection point $\textit{I}_n$ of the line $L_{n}$ with the bottom axis of the image; we only retain lines that intersect within the image bounds. 
Each line is then described by its point of intersection $I_{n}=[x, y]$ and the angle of intersection~$\phi_{n}$ such that~$L_{n}=(I_{n}, \phi_{n})$.


We use the estimated crop lines in conjunction with the moving variance~\cite{macgregor1993exponentially} to represent the local structure of the field.
%
The moving variance of the estimated crop line angles, $\phi_{n}$, is calculated over a window of size $k$ such that
\begin{equation}
    \sigma^2(\phi_n) = \dfrac{\sum_{i=n-k/2}^{n+k/2} (\phi_n - \bar \phi_n)^2}{k}; \;
    \bar \phi_n = \frac{\sum_{i=n-k/2}^{n+k/2} \phi_i}{k}.
    \label{eq:movar}
\end{equation}
The moving variance operator yields peaks when there is discord between the local hypothesised crop lines, this occurs between the crop-row lines.
Troughs occur when there is consistent agreement regarding the hypothesised crop lines, this occurs in the presence of crop-rows.
A depiction of this field structure signal is given in~\figref{fig:movar_graph}.

The peaks ($\blacktriangle$) and troughs ($\triangledown$) of the field structure signal are detected using peak prominence with a constant threshold.
To detect the troughs, the signal is flipped (via negation) and peak prominence is applied with the same threshold. 
The detection of troughs is more complex as crop-rows can yield multiple peaks. 
We resolve this by computing the weighted average of the possible solutions in the local neighbourhood, where the local neighbourhood is defined to be adjacent sampling positions with similar standard deviation values.
An example of this is given in~\figref{fig:movar_graph} where the final trough is denoted by $\blacksquare$. 
The output of this step is the set of detected crop-row lines $\mathcal{L}$.

\subsubsection{Crop-Row Tracking}
\label{sec:tracking_crop_rows}
For each detected crop-row line~$L_{n}$ we form a parallelogram~$\mathcal{P}_{n}$ with a width relative to the canopy width of the crops forming line~$L_{n}$.
All parallelograms have a minimum constant width of $5\%$ of the image size to avoid shrinking to zero size width in regions without vegetation or having only tiny plants.
For every new camera image, we update the position of each line~$L_{n}$ based on the crop centers within the parallelogram~$\mathcal{P}_{n}$ at its latest position.
If a crop-row line is not detected, the position of the previous line (and parallelogram) is maintained for a few seconds, this allows our approach to continue guiding the robot even when there is an uneven crop distribution in the crop-rows.
The $\mathcal{L}$ is then tracked and used to update the robot's position to ensure that the crop is always beneath the robot, using a visual-servoing controller. 
Finally, when all of the crop centers are in the top half of the image we consider that we have reached the end of the crop-row and employ the multi-crop-row switching algorithm (\secref{subsubsec:switching}).

%

%% file: contribs/vs.tex
   
To guide the robot within the crop-rows, we utilize our previously developed in-row guidance approach~\cite{ahmadi2020visual} which relies on the image-based visual servoing (IBVS) controller~\cite{espiau1992new}.
The guiding feature $\mathcal{\mathbf{L}}$ is the average line of the set of detected crop-row lines $\mathcal{L}$ illustrated in \secref{subsec:crop_row_extraction} computed from the current camera image.
The desired feature~$\mathcal{\mathbf{L}}^{*} = [0,\,\frac{H}{2},\,0]$ is located at the bottom  center  of  the  image~$\mathcal{I}$, illustrated in \figref{fig:feature_space}.
And, the IBVS controls law leads the current feature $\mathcal{\mathbf{L}} = [a,\,\dfrac{H}{2},\,\Theta]$ to towards $\mathcal{\mathbf{L}}^*$, where $H$ denotes the height of the image and $a$ is the deviation from  the  image center of the intersection point $\mathbf{I}$.  
By continuously regulating the robot's motion based on this controller and observation features explained in \secref{sec:tracking_crop_rows}, we ensure the robot stays in the middle of the path and follows the desired crop-rows to their end.
At the end of crop-rows, we switch to the next lane by using the switching technique described in the next section.

%% file: contribs/switching.tex
To autonomously navigate over an entire row-crop field a robot must be able to both navigate down lanes and shift between them.
Utilizing an only image-based motion controller in conjunction with other localization techniques (like GPS and wheel odometry) could considerably improve the reliability of the system in cases of outage of motion information due to hardware problems and environmental situations.
In our previous work~\cite{ahmadi2020visual} the task of changing lanes was managed successfully, however, it was only designed to handle a single crop-row in a lane under a highly engineered condition.
This method was strongly reliant on the seeding pattern of the crops-rows and struggled with cases which often occur in a real-fields like: uneven seeded crop-rows, unexpected distances between the rows, and appearance cluttered and weedy regions.
Furthermore, it required significant space to perform the switching maneuver and could not reconcile differences between rows and subsequently following the incorrect lane. 

In this work we propose a multi-crop-row switching technique that detects and counts the rows as it progressively shifts between them. 
This takes advantage of the side-ways movement of BonnBot-I allowing easier transitions while requiring less space.
To detect a new crop-row we exploit SIFT features to ensure we only traverse the desired number of crop-rows to confirm our new lane is in the correct location without relining on any motion information neither odometry nor GPS.
\begin{figure}[t!]
	\centering
	\vspace{3mm}
	\includegraphics[height=4.5cm]{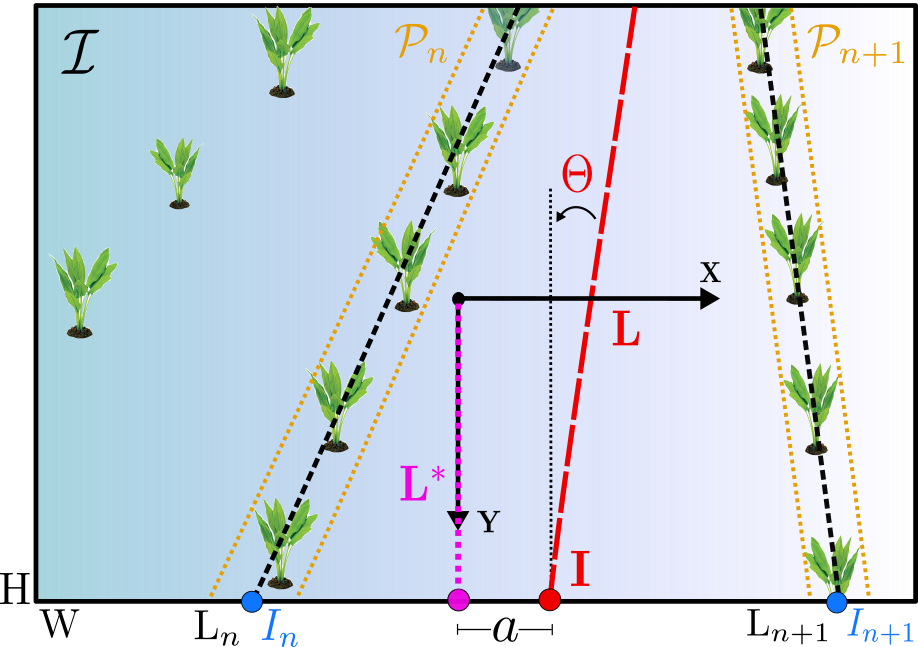}
	\caption{The image frame $\mathcal{I}$ and $\mathbf{L} = [\mathbf{I},\,\Theta]$ denotes the dominant line estimated from the visible crop-rows.}
	\label{fig:feature_space}
	\vspace{-6mm}
\end{figure}


We start by considering the robot to have found the end of the crop-row; stage $\circled{3}$ of the navigation scheme~\secref{subsec:guidanceStrategy}.
The multi-crop-row detection algorithm~\secref{subsec:crop_row_extraction} provides us with the number of crop-rows $C$ that have been traversed.
To find new crop-rows, lane switching, we need to move across $C$ rows and then restart the crop-row following algorithm.
To do this we describe each crop-row by a set of SIFT features and follow the algorithm described below. 


Assuming we are moving left-to-right, we store the features of the right most parallelogram in the image forming a feature set $\mathcal{G}$.
The robot then starts moving to the right-side with a constant $u_y = 0.15 \, m/s$ velocity.
Upon receiving a new image we detect the crop-rows in a similar manner outlined in~\secref{subsec:crop_row_detection} and then only consider the right most side of the image.
We extract a new set of SIFT features from the right most parallelogram in the image forming a feature set $\mathcal{G}^{*}$.
The new feature set, $\mathcal{G}^{*}$, is potentially a new crop-row. 
To determine if $\mathcal{G}^{*}$ is a new crop-row we compare it to the stored SIFT features $\mathcal{G}$.
If a new crop-row has been detected, we update the stored features ($\mathcal{G}$=$\mathcal{G}^{*}$) and continue this process until we have moved across $C$ new crop-rows.
\begin{figure}[t!]
    \vspace{2mm} 
    \begin{subfigure}[b]{28mm}
        \includegraphics[height=20mm,width=28mm]{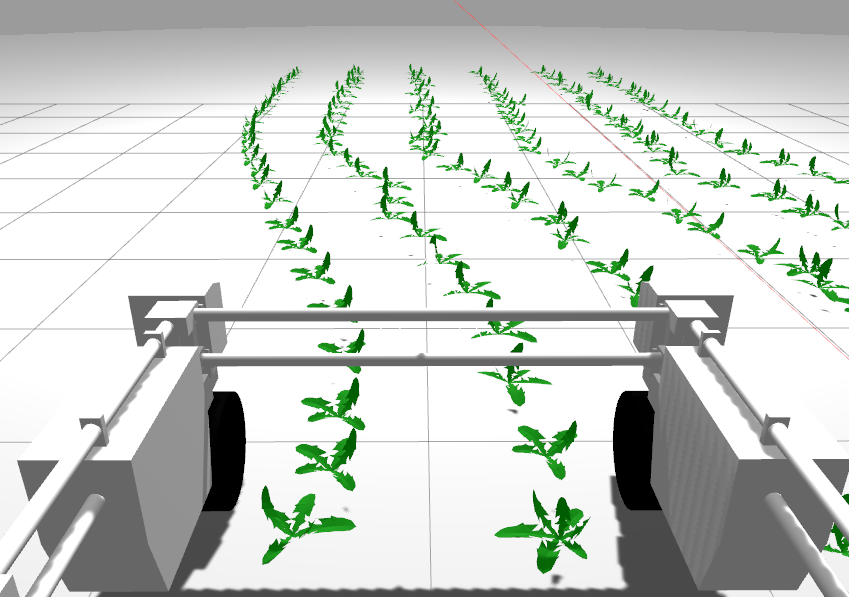}
        \caption{Sim-Curved}
        \label{fig:lemon_balm_graphic}
    \end{subfigure}    
    \begin{subfigure}[b]{28mm}
        \includegraphics[height=20mm,width=28mm]{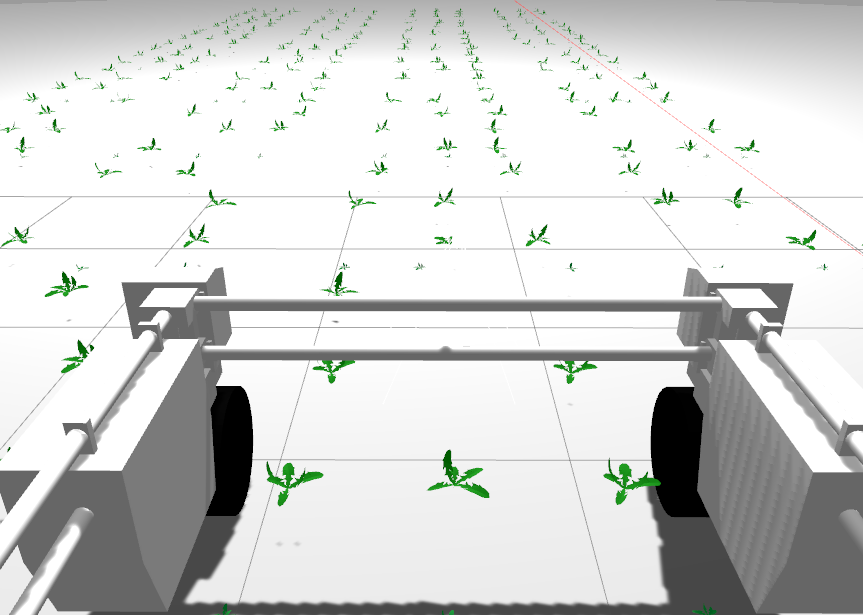}
        \caption{Sim-Large-Gaps}
        \label{fig:coriander_graphics}
    \end{subfigure} 
    \begin{subfigure}[b]{28mm}
        \includegraphics[height=20mm,width=28mm]{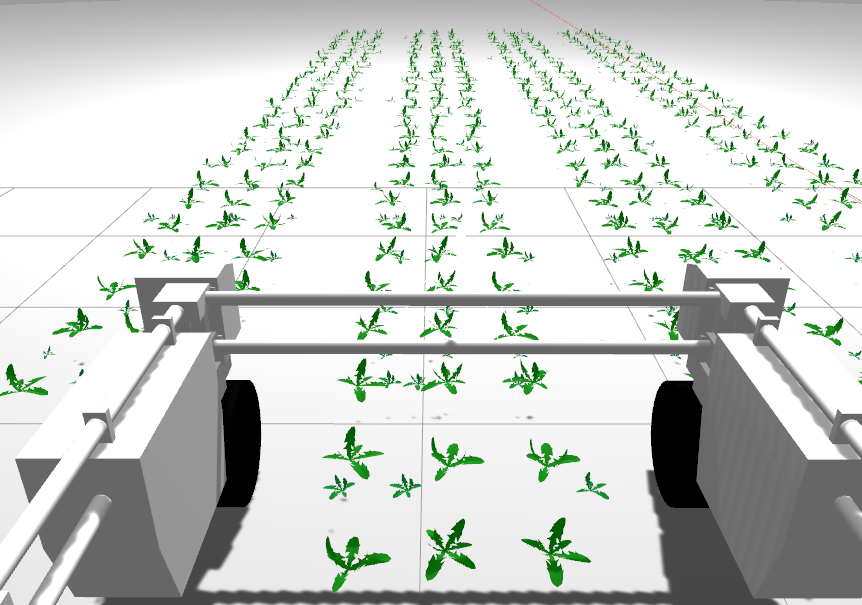}
        \caption{Sim-Dense-Weed}
        \label{fig:coriander_graphics}
    \end{subfigure} 
    
    \caption{Simulated Fields with different plant sizes as weeds (small) and crops (big), for purpose of visualization soil background in simulation is removed.}
    \label{fig:simulatedCropRows}
    \vspace{-6mm} 
    
\end{figure} 
\begin{figure*}[!t]
    \centering
    \vspace{3mm}
    \hspace{1pt}
    \includegraphics[width=43mm]{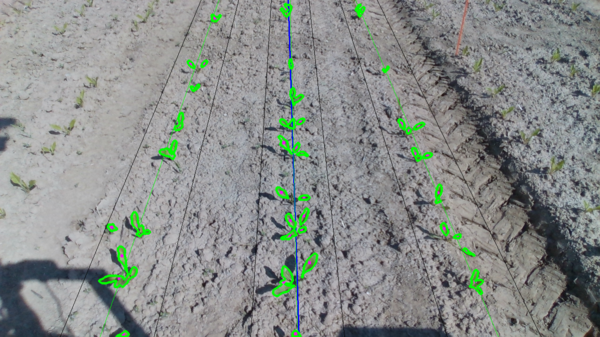}
    \includegraphics[width=43mm]{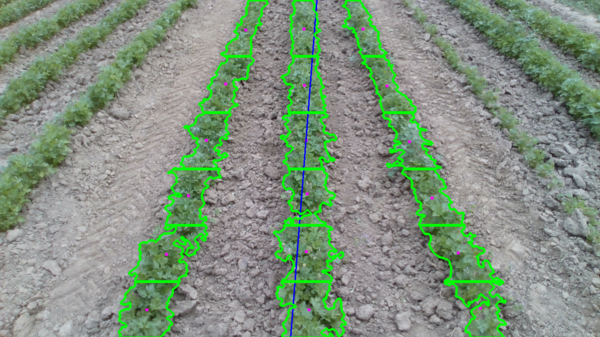}
    \includegraphics[width=43mm]{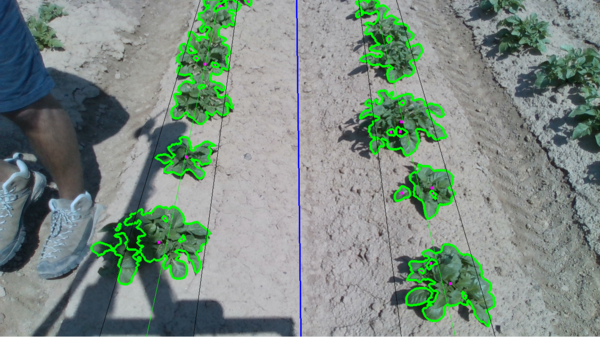}
    \includegraphics[width=43mm]{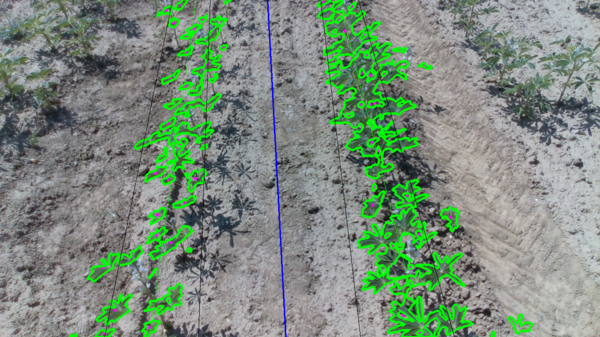} \\
    \vspace{3pt}
    \begin{subfigure}[b]{43mm}
         \includegraphics[width=43mm]{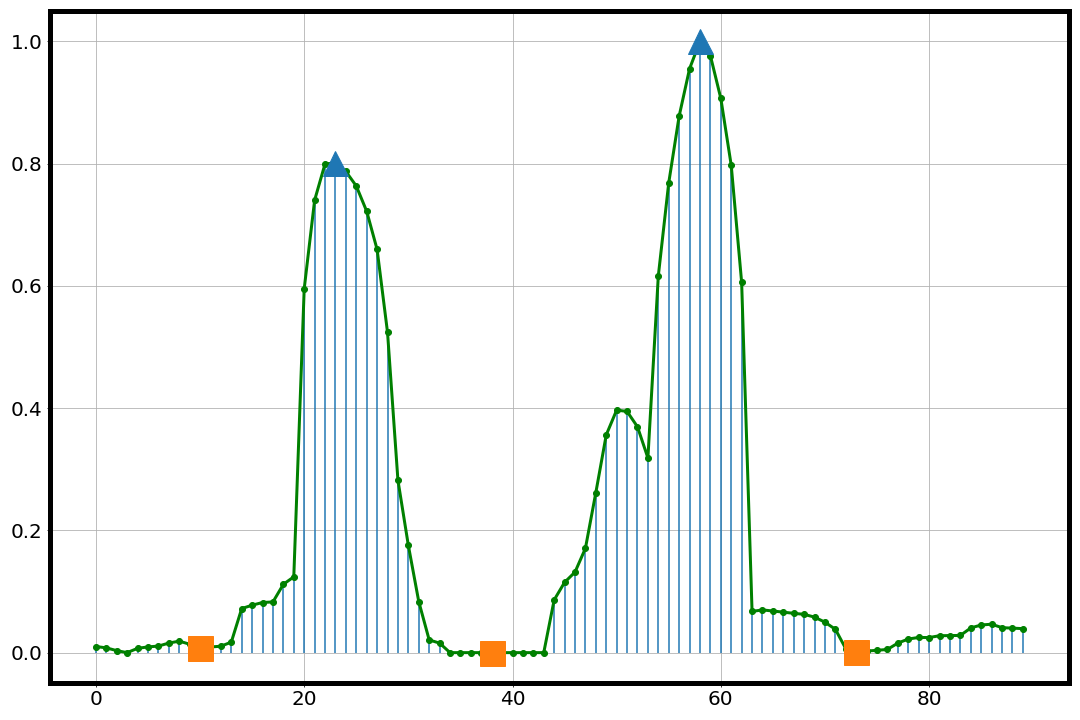}
        \caption{Sugar-beet}
        \label{fig:movar_sb}
    \end{subfigure}
    \begin{subfigure}[b]{43mm}
        \includegraphics[width=43mm]{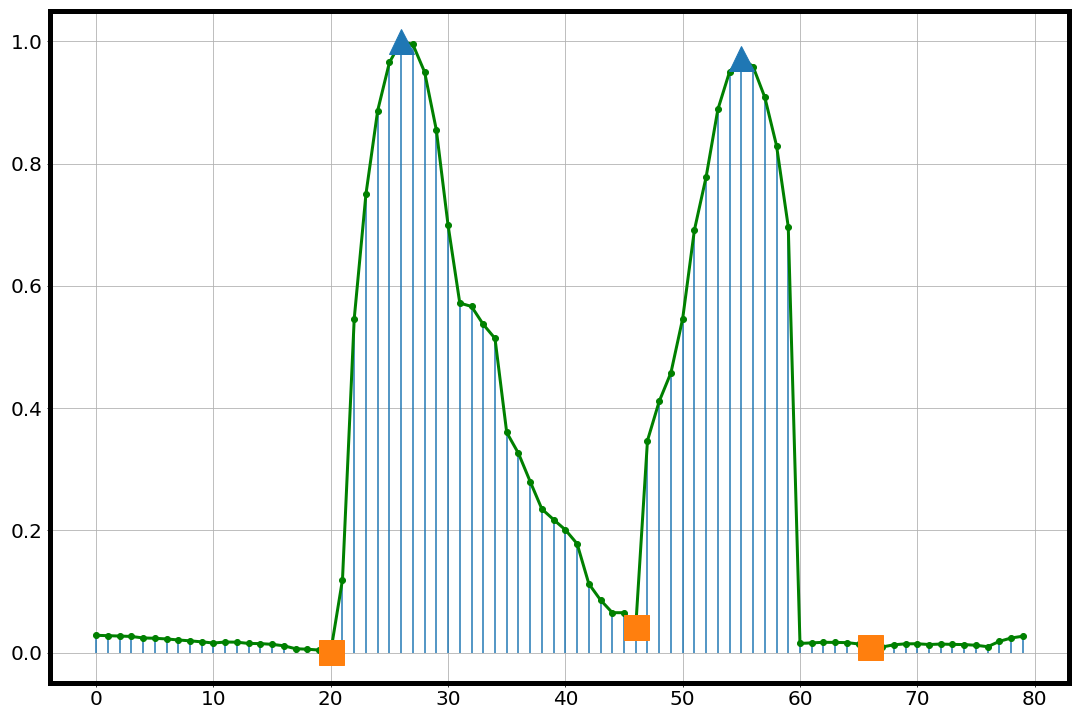}
        \caption{Coriander}
        \label{fig:movar_coriander}
    \end{subfigure}
    \begin{subfigure}[b]{43mm}
        \includegraphics[width=43mm]{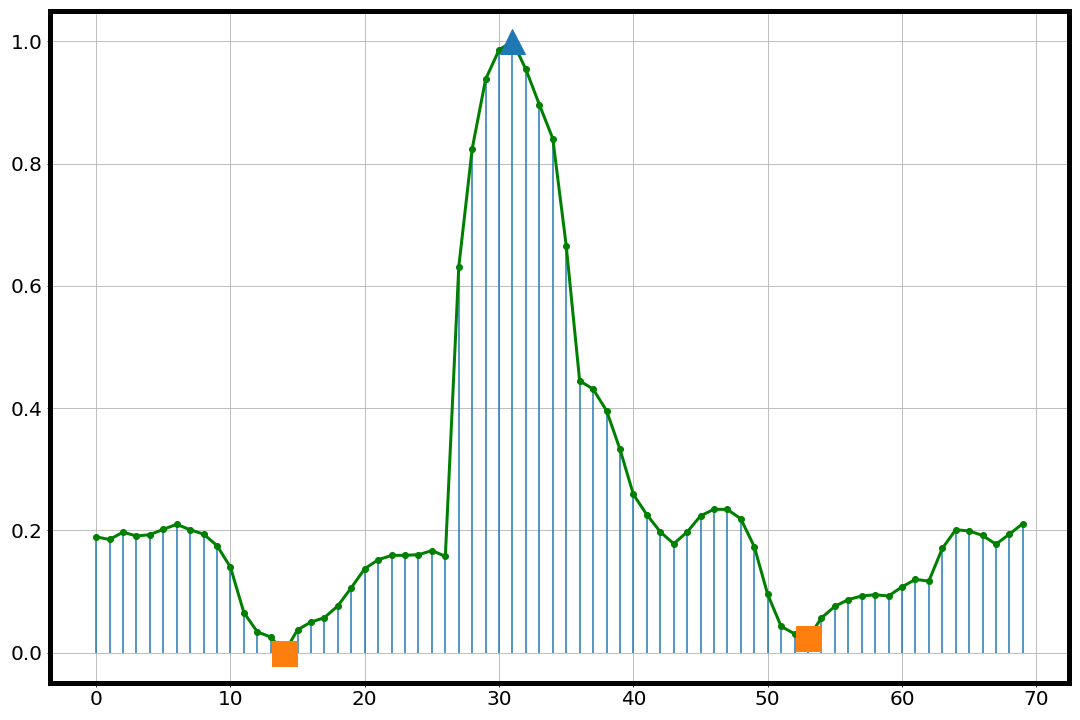}
        \caption{Potato}
        \label{fig:movar_potato}
    \end{subfigure} 
    \begin{subfigure}[b]{43mm}
        \includegraphics[width=43mm]{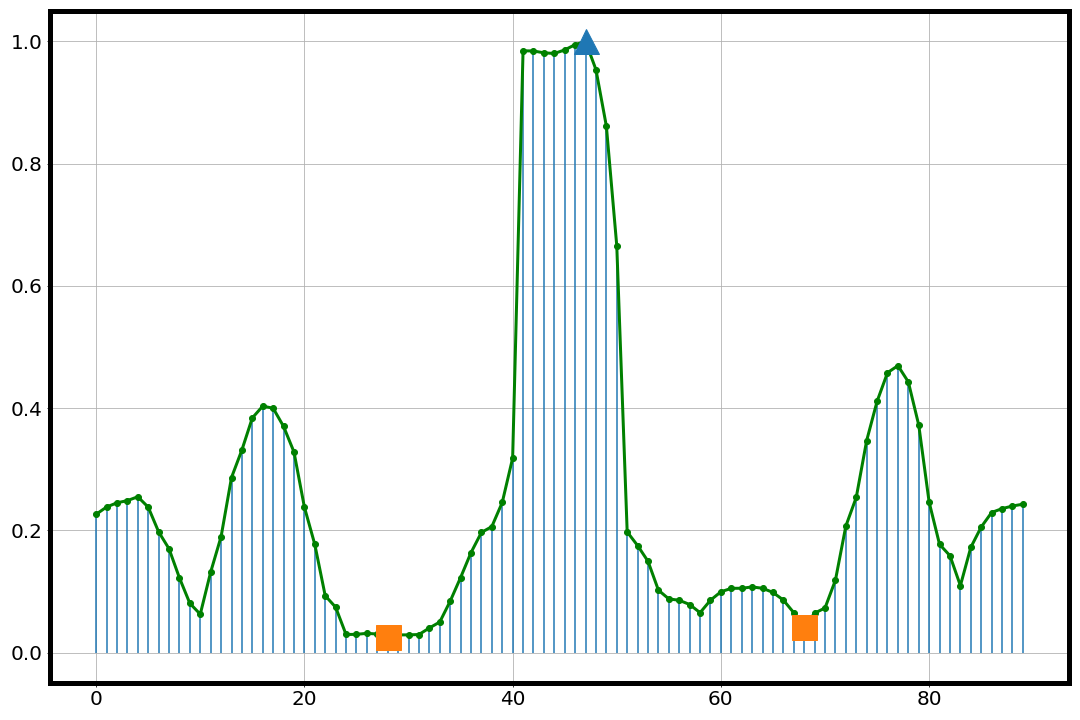}
        \caption{Beans}
        \label{fig:movar_beans}
    \end{subfigure}
    \caption{Illustrations of four crops (top row) RGB images with marked vegetation index of detected rows and their corresponding field structure signal (bottom row).
    The detected peaks (blue triangle) and troughs (orange square) obtained via their prominence in the signal are also provided.
    Denote that, the field structure signal only include the values of lines intersecting with the bottom axes of the image.}
    \vspace{-6mm}
    \label{fig:fields}
\end{figure*}

To compare the SIFT features, we use a FLANN-matcher to obtain the best matches $\mathbf{\Omega}$ between the two sets $\mathcal{G}$ and $\mathcal{G}^{*}$.
This results in the Euclidean distances between $\mathcal{G}$ and $\mathcal{G}^{*}$ being stored in $\mathbf{\Omega}$. 
We then take the average of the $m$ matches in $\mathbf{\Omega}$ which are above a threshold $\lambda$.
This is used to provide a distance measure between the two sets of features:
\begin{equation}
    \vspace{-2mm}
    \label{eq:similarity_distance}
    D(\mathcal{G}, \mathcal{G}^{*}) = \dfrac{1}{m} \sum_{i=1}^m \mathbf{\Omega}_m
\end{equation}
%
\noindent When $D(\mathcal{G}, \mathcal{G}^{*})$ exceeds a threshold $\tau$ we assume a new crop-row has been found, $\tau$ is a crop type specific constant.


%% file: contribs/exp.tex
We performed three experiments to show the capability and robustness of our proposed  approaches. 
These experiments were carried out on both simulated and real phenotpying fields.
The simulated fields are built in the Gazebo environment with either two or three rows in a lane.
It is also designed with various challenging arrangements, such as: curved crop-rows (Sim-Curved), crop-rows with large inter-crop gaps (Sim-Large-Gaps), and lanes of crops having dense weed appearance (Sim-Dense-Weed), which are depicted in~\figref{fig:simulatedCropRows}.
The real fields represent up to five different crops with non-similar canopy shapes and varying numbers of crop-rows per lane. 
All results outlined in this section are based on the evaluation data and required limited human intervention (apart from minor hyper-parameter tuning) during run-time and navigation.
%
%
\subsection{Experimental Setup}
The experiments of all five crop types were completed on BonnBot-I at Campus Klein-Altendorf of the University of Bonn under various illumination and weather conditions such as: relatively wet and very dry grounds, cloudy and sunny day-times with long and short shadow cases (with minimal hyper-parameter tuning).
To fast track the field experiments we used a 1:1 scale Gazebo simulation model with a realistic field. 
From this simulation, we were able to derive our algorithmic hyper-parameters.
On BonnBot-I, the front and back navigation cameras are fixed at a height of $1.0 \, m$ and tilt angle $\rho = 65^{\circ}$. 
Both camera resolutions are $1280\times720$ with a capture rate of $15\,fps$.
For all experiments, the width $w$ of sliding window $\Psi$ was kept constant $w=W/10=128$ with a height of $h=720$ pixels.
This window size and $\mathbf{S}=13$ ensures $\simeq95\%$ overlap between consecutive sliding windows.
Also, we empirically set $k=10$ in \eqref{eq:movar} which in simulation provided the best trade-off between sample consistency and neighbourhood relationship. 
As the primary goal of this platform is to perform weeding we set its velocity to be a constant $v_x^{*}= 0.5 \, m/s$.
We use differential velocity control within the crop-rows and omni-directional control for switching between the lanes.
Our approach is implemented using Python and PyCuda ensuring real-time operation (while with CPU only machines performance is still acceptable) and runs on a user grade computer~(Cincose DS-1202).

\subsection{Multi-Crop-Row Detection}
\label{subsec:exp_autotune}

The first experiment is a qualitative analysis on the ability to detect crop-rows in the field using the technique described in~\secref{subsec:crop_row_extraction}.
We use ExG with Otsu's technique~\cite{reimann2005background} to differentiate foreground from background as this obviates the need for individual hand-tuned plant segmentation thresholds.
The goal of this technique is to exploit the dominant crop locations and accurately detect the best location for traversing a lane (i.e.~keeping the crop-rows under the platform).
Due to weeds growing between the crop-rows this can be a challenging proposition in real fields.
Qualitative results for four crops (sugar-beet, coriander, potato, and beans) are presented in~\figref{fig:fields}; we refrained from adding Lemon-balm and simulated fields results due to space limitations.
The illustrated crops have diverse canopy types (see~\figref{fig:fields}) and are arranged in two standard patterns with two and three crop-rows per lane.
In the bottom row of~\figref{fig:fields} it can be seen that our approach to detection crop-row lines through peaks and troughs works for these chosen crops even with varying number of crop-rows.
This is true even for the challenging crops such as coriander and beans.
\begin{figure}[b!]
    \vspace{-5mm}
	\centering
	\includegraphics[width=\linewidth]{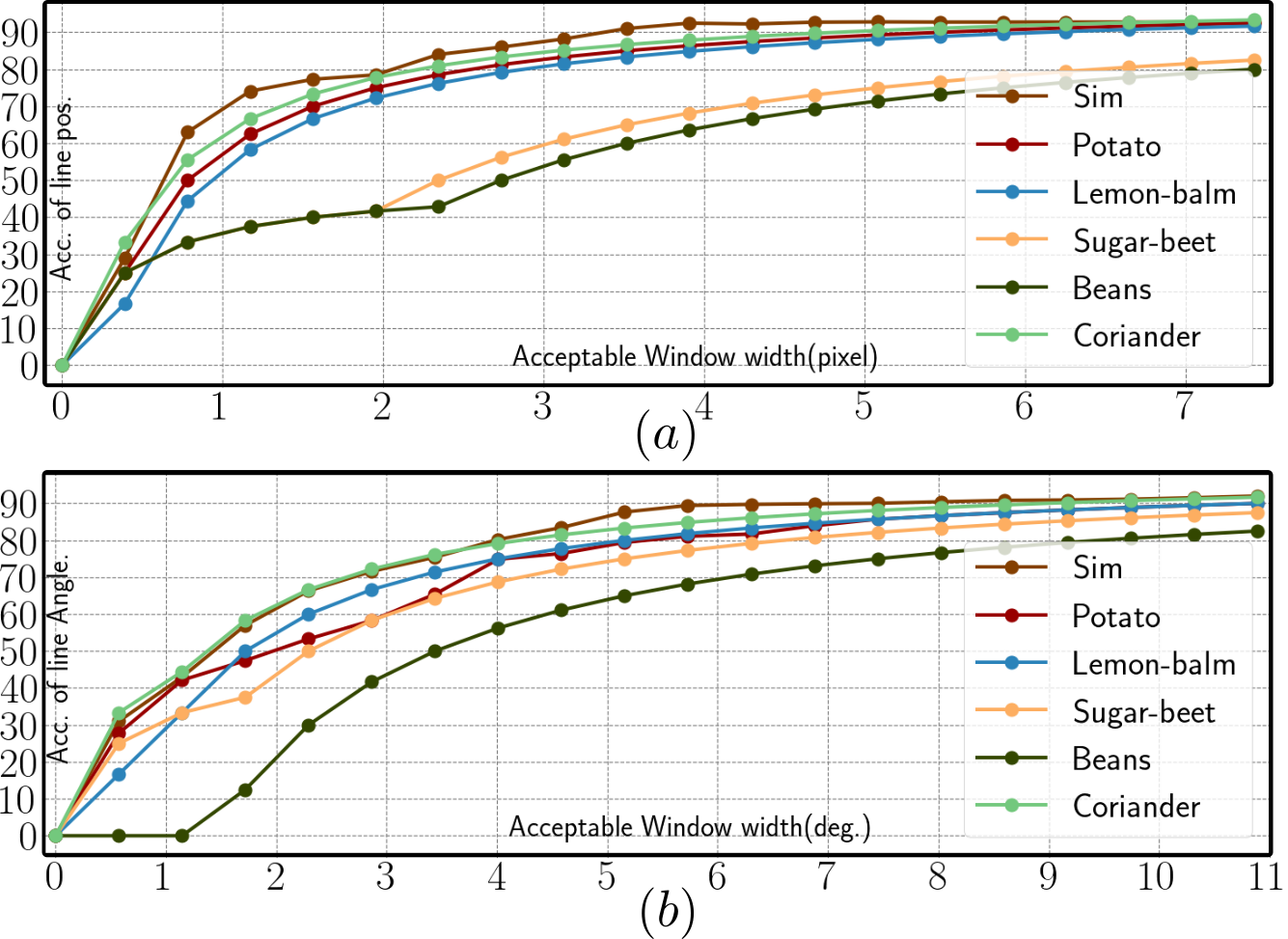} \\
	\caption{Performance of multi crop-row detection technique: accuracy of detections (a) position (w.r.t the image width) and (b) orientation w.r.t the acceptable thresholds}.
	\label{fig:lineDetAcc}
\end{figure}
%

For all crops we were able to consistently detect both the peak (no crop-row) and trough (crop-row) locations regardless of the presence of weeds.
This is especially evident in coriander where even with the small distance within the crop-row (between the coriander plants) we are still able to detect the crop-rows.
This is an example where dividing single large regions into sub-regions is essential.
Sugar-beet,~\figref{fig:fields}-a, is another interesting use case. 
Visually it is considerably more difficult to discern the crop locations, however, this technique was still able to extract the required locations (crop-rows).
Overall, this technique for crop-row detection successfully located the required troughs in order to navigate a lane, providing accurate information required for the other stages of our system. 

	
	
    

To further analyze the robustness of crop-row detection technique we perform a quantitative evaluate.
For each of the five crops (those listed in \figref{fig:fields} and lemon-balm) and the simulated field, 100 images were annotated using data from BonnBot-I where the camera tilt angle $\rho$ was varied from $55^{\circ}$ to $75^{\circ}$.
The annotations contain ground-truth of all the crop-row lines located underneath the robot and crop masks belonging to the main crop-rows.
To measure the accuracy we compare the predicted lines of each image to the ground-truth using two parameters: position and orientation.
The position of a line is defined based on its intersection with bottom edge of the image, where the distance between the prediction and the ground truth is normalized based on the width of the image.
\figref{fig:lineDetAcc} outlines the quantitative performance result of real fields and averaged performance of simulated fields.

We observe our method is able to estimate crop row positions with a mean accuracy of $88.1\%$ and standard deviation of $8\%$ over all types of real crops.
Similarly, we see that the algorithm is able to correctly estimate the orientation of crop-rows in more than $88.3\%$ of cases when the acceptance threshold is set to $11$ degrees.
The crop-row lines of beans and sugar-beet were the hardest to estimate.
For sugar-beet, we attribute this to the fact that the crop was at an early growth stage, as seen in \figref{fig:fields}, and this made it more complicated to detect the crop lines.
For beans, we attribute this to its branchy canopy shape, disarranged seeding pattern, and plant vibration due to the wind in the field.
One potential use case where the approach may fail is when the number of weeds is close to or greater than the number of crops in the image (very high weed pressure).
Furthermore, difficulties may be faced when navigating the field once full canopy closure has been achieved and there are no visible crop lanes to follow (full vegetation). Nevertheless, overall we observe our novel crop-row detection method could estimate lines of crops in a variety of challenging real-world conditions for different crop types reliably.

\subsection{Navigating Along The Crop-Rows}
\label{subsec:exp_navigation}

To analyse the performance of our crop-row navigation technique we require accurate ground truth information.
We collected the ground truth information by manually driving the robot down each of the row-crop fields for all crop types and stored the associated information (e.g. GPS measurements) for later evaluation.
Also, all simulated crop rows came with reference lines coordinates from simulation environment.
The associated GPS measurements are then used as the ``correct'' position (accurate to $1cm$).
Even though manual operation can cause some errors we consider this to be an appropriate ground truth to compare to as the crop-rows are not guaranteed to be planted in a straight line.

The five crops (sugar-beet, coriander, potato, beans, and lemon-balm) provide a range of challenges such as different canopy types, weed densities, and varying growth stages.
\tabref{tab:following_results} outlines the performance of our full pipeline, including the navigation system, on these crops as well as three challenging simulated fields.
The most challenging crop for navigation was sugar-beet and we attribute this to two reasons. 
First, the crop was at an early growth stage, as seen in \figref{fig:fields}, and this made it more complicated to detect the crop lines. 
Second, not all of the sugar-beet had germinated and this led to gaps or long ``dead space'' along the rows, which the same effect can be seen in Sim-Large-Gaps results too. 
However, tracking multiple crop-rows allowed our technique to still navigate over the entire evaluation area without any manual intervention.
This evaluation shows that multiple crop-row following has considerable benefits over techniques that only track a single crop-row.
  
\begin{table}[!b]
	\centering  
    \vspace{-6mm}
	\caption{Lane following performance of BonnBot-I using proposed method in real and simulated fields.}

	\begin{tabular}{c c c c}
	\toprule
	
	\textbf{Crop}
	& \textbf{Length}
    & \begin{tabular}[c]{@{}c@{}}\textbf{$\mu \pm \sigma$ of}\\ \textbf{dist. to crop-rows}\end{tabular}
	& \begin{tabular}[c]{@{}c@{}}\textbf{$\mu \pm \sigma$ of}\\ \textbf{angular error}\end{tabular}\\\hline
	\midrule 
    \textit{Sim-Curved}     & 200\,m & 9.01 $\pm$ 2.63\,cm & 4.52 $\pm$ 3.52\,deg \\
    \textit{Sim-Large-Gaps} & 200\,m & 6.75 $\pm$ 3.15\,cm & 4.76 $\pm$ 2.69\,deg \\
    \textit{Sim-Dense-Weed} & 200\,m & 7.41 $\pm$ 2.86\,cm & 3.91 $\pm$ 1.73\,deg \\
    \midrule
    \textit{Beans}          & 52\,m & 3.49 $\pm$ 2.89\,cm & 3.73
    $\pm$ 3.21\,deg \\
    \textit{Potato}         & 37\,m & 2.18 $\pm$ 3.01\,cm & 4.91 $\pm$ 1.63\,deg \\
    \textit{Coriander}      & 54\,m & 2.91 $\pm$ 2.38\,cm & 2.57 $\pm$ 1.05\,deg \\
    \textit{Sugar-beet}     & 69\,m & 8.41 $\pm$ 3.79\,cm & 3.25 $\pm$ 1.27\,deg \\
    \textit{Lemon-balm}     & 40\,m & 2.12 $\pm$ 1.58\,cm & 3.21 $\pm$ 2.83\,deg \\

        \bottomrule
    \end{tabular}

	\label{tab:following_results}
\end{table} 

From a navigational perspective the bean crop had a large standard deviation between real fields and Sim-Curved among simulated fields when considering angular error.
The weather conditions played a crucial part in this as heavy winds consistently changed the location of the leaves of the crops.
This limitation in the navigation technique leads to large angular variations while traversing the lane.

Across the five real crop types the average deviation from the ground truth was $3.82cm$ or approximately $10\%$ of the crop-row distance.
This minor fluctuation is sufficient to ensure safe navigation without damaging crops. 
Finally, this navigational accuracy was sufficient for the technique to traverse all the crops in the field without manual intervention.

\subsection{Multi-Crop-Row Switching}
\label{subsec:exp_switching}

\begin{figure}[t!]
	\centering
	\vspace{2mm}
	\includegraphics[width=0.8\linewidth]{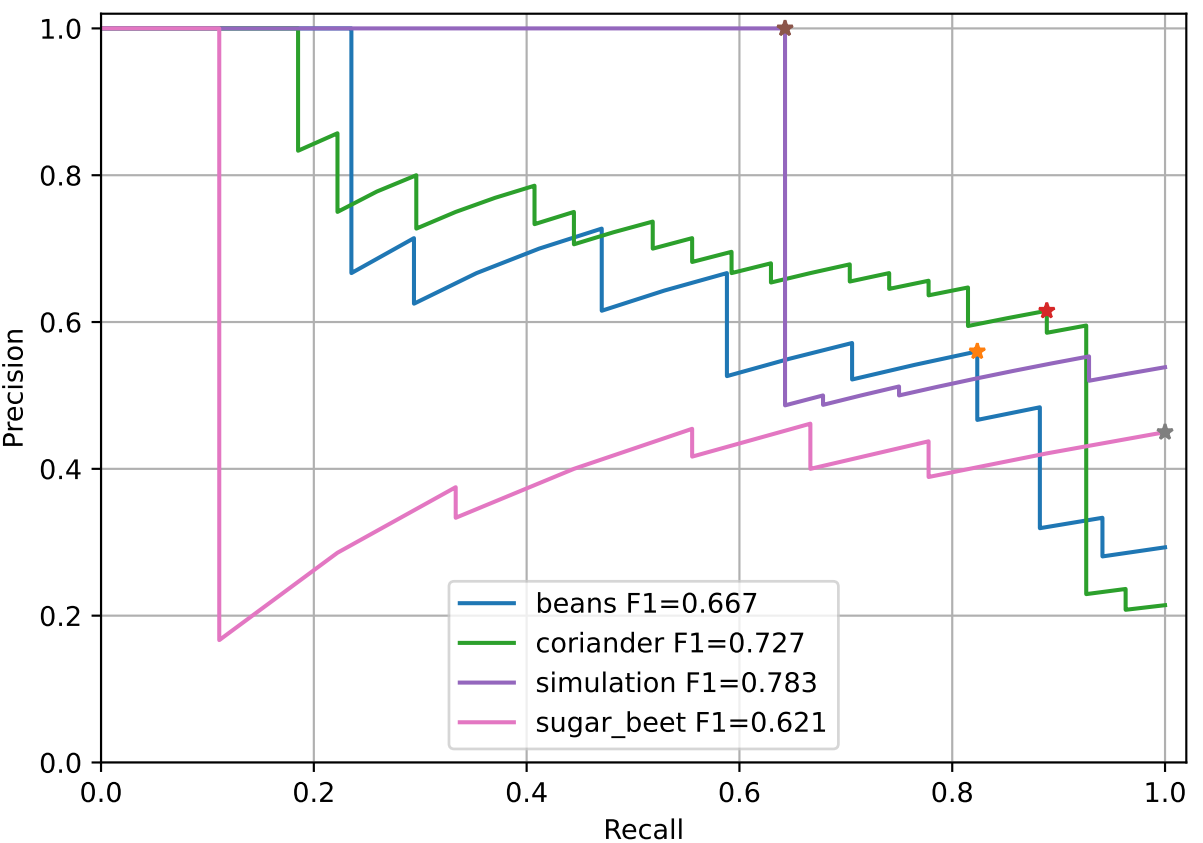}
	\caption{The precision recall plot for switching the platform across lanes, includes, beans, coriander, and sugar-beet.}
	\label{fig:switchpr}
	\vspace{-6mm}
\end{figure}

Our final evaluation is based on the lane switching technique outlined in~\secref{subsubsec:switching}.
To evaluate the performance of this technique we manually annotated randomly selected positive and negative samples from our three main crop types: beans, coriander, and sugar-beet and simulated fields as we did not have the switching information for potato and lemon-balm due to a technical problem.
We store one of the positive annotations as our main row and compare it to each of the other positive and negative rows.
\figref{fig:switchpr} outlines the precision-recall curves achieved on each of the main crops.

We outline the F1 score here which provides a trade-off between precision and recall.
For this simple matching technique we are able to achieve promising results across all crop types, even sugar-beet which, like outlined \secref{subsec:exp_navigation}, had a number of added complexities.
The early germination stage of the crop added extra complications to crop-row switching as, even visually, the rows appeared similar.
However, we were able to achieve an F1 score of 62.1, which was the lowest performing crop. 

Overall, from these evaluations we were able to empirically set thresholds that favored high precision in order to remove false positives.
From this we were able to provide a lane switching technique that was robust to the challenges of each crop type.
In experiments in the field and deployed on a robot, it successfully switched 6 lanes of crop, across the three main crop types, without any manual intervention.
Furthermore, we used this technique in simulated crop-row fields for $20$ lane switching cases which outlined an average success rate of $90\%$ percent.

A final key analysis of our crop-row switching technique is the distance needed to perform the maneuver.
In our experiments an average of $0.7m$ was required from the end of the crop to the location of the camera.
This is a marked improvement over~\cite{ahmadi2020visual} which required more empty space to perform the switching than the length of the robot itself.

